\documentclass[10pt,a4paper]{article}
\usepackage{graphicx}
\usepackage{subfigure}
\usepackage{amsmath,amsthm}
\usepackage{amssymb,amsfonts}
\usepackage{authblk}
\usepackage{url}
\usepackage{hyperref}
\usepackage{multirow}
\usepackage{subfigure}
\usepackage{booktabs}

\newtheorem{definition}{\textbf{Definition}}

\newtheorem{theorem}{\textbf{Theorem}}

\newtheorem{corollary}{\textbf{Corollary}}

\title{Photometric Redshifts with Copula Entropy}
\author{Jian MA\thanks{Email: majian@hitachi.cn}}
\affil{Hitachi China Research Laboratory}
\date{}

\begin{document}
\maketitle

\begin{abstract}
In this paper we propose to apply copula entropy (CE) to photometric redshifts. CE is used to measure the correlations between photometric measurements and redshifts and then the measurements associated with high CEs are selected for predicting redshifts. We verified the proposed method on the SDSS quasar data. Experimental results show that the accuracy of photometric redshifts is improved with the selected measurements compared to the results with all the measurements used in the experiments, especially for the samples with high redshifts. The measurements selected with CE include luminosity magnitude, the brightness in ultraviolet band with standard deviation, and the brightness of the other four bands. Since CE is a rigorously defined mathematical concept, the models such derived is interpretable.
\end{abstract}

\section{Introduction}
Photometric redshifts is a key problem in all-sky extragalactic surveys and can be applied to studying galaxy evolution and cosmology \cite{Newman2022}. It is a inexpensive way of probing the universe than spectroscopy, which makes it a good means for obtaining distance information of billions of objects in our expanding universe. Spectroscopy could be followed up for more precise observations if some objects are selected through redshifts estimation. Though widely adopted, photometric redshifts has its own drawbacks for the performance of redshift estimation.

Accuracy is a big performance issue in photometric redshifts \cite{Salvato2019}. One of the sources of prediction bias is the machine learning models ill-built with all the photometric measurements. Photometric redshifts is essentially a regression problem that predicts redshifts of objects from their multi-band photometric measurements. Many machine learning methods, such as SVM \cite{Zheng2012}, Gaussian processes \cite{Miller2015}, quantile regression \cite{Meshcheryakov2018}, among many others \cite{Zhang2020,Newman2022}, have been applied to this problem. A review paper on applications of machine learning to photometric redshift is given by Zheng and Zhang \cite{Zheng2012a}. In the past research, all the photometric measurements were used for building regression models. It is problematic because only a subset of measurements may be useful for the redshifts estimates. It is possible to improve the accuracy of estimates by selecting those useful measurements out.

Variable selection is a common problem in statistics and machine learning. It selects only a subset from all the available variables to build the regression models. In this way, the accuracy of the predictive models can be improved with the selected variables. Variable selection can be based on likelihoods, such as AIC, BIC, or accuracy, such as LASSO \cite{Tibshirani1996}, or correlation, such as HSIC \cite{Gretton2007}, distance correlation \cite{Szekely2007}, and copula entropy \cite{Ma2021a}.

Copula Entropy (CE) is a recently proposed mathematical concept for measuring multivariate statistical independence \cite{Ma2011}. It is proved to be equivalent to mutual information in information theory. Compared with traditional correlation measures, such as the Pearson correlation coefficient, which can only be applied to linear cases with Gaussianity assumption, CE can be applied to any case without any assumption on the distribution of random variables. A non-parametric method for estimating copula entropy has also been proposed based on rank statistic \cite{Ma2011}. CE-based variable selection has been proposed \cite{Ma2021a} and applied in many scientific fields, such as hydrology \cite{Chen2013}, medicine \cite{Mesiar2021}, among others.

In this paper we propose to apply copula entropy to photometric redshifts. Particularly, the CEs between redshifts and multi-bands photometric measurements will be estimated from real data, and then those associated with large CEs will be used for predicting redshifts. 

CE has several merits for photometric redshifts. It is model-free and hence can be applied directly to measure the nonlinear relationships between photometric measurements and redshifts without making any assumptions on the underlying problems. It is a rigorously defined mathematical concept, so we are confident in its applications to any relationships in cosmology and enjoy its advantages over many other similar methods, such as random forests \cite{Breiman2001}, that can also be used to derive variable importance for regression models but without theoretical support. The correlation relationship derived with copula entropy is also interpretable, which is important in general \cite{Huppenkothen2023} and here for understanding which measurements are more important and how important they are for redshifts estimation and therefore provide evidence for further astrophysical study and instrument design. 

We applied our method to the Sload Digital Sky Survey (SDSS) quasar catalog data \cite{Schneider2007} to verify the effectiveness of our method. A subset of photometric measurements were selected out of all the five-bands measurements, including luminosity, ultraviolet-bands brightness with our method. Experimental results show that using these measurements selected with CE as inputs of SVM and random forests can lead to more accurate redshift estimates, especially on high redshifts ($z>4.0$), than using all the measurements.

This paper is organized as follows: Section \ref{s:method} introduces copula entropy and machine learning algorithms used in this study; Section \ref{s:exp} presents experiments and results; Section \ref{sec:discussion} gives some discussion;  and finally Section \ref{sec:con} concludes the paper.

\section{Methodology}
\label{s:method}
\subsection{Copula Entropy}
\label{s:CopEnt}
\subsubsection{Theory}
Copula theory is about the representation of multivariate dependence with copula function \cite{joe2014,nelsen2007}. At the core of copula theory is Sklar theorem \cite{sklar1959} which states that multivariate probability density function can be represented as a product of its marginals and copula density function which represents dependence structure among random variables. Such representation separates dependence structure, i.e., copula function, with the properties of individual variables -- marginals, which make it possible to deal with dependence structure only regardless of joint distribution and marginal distribution. This section is to define an statistical independence measure with copula. For clarity, please refer to \cite{Ma2011} for notations.

With copula density, Copula Entropy is define as follows \cite{Ma2011}:
\begin{definition}[Copula Entropy]
	\label{d:ce}
	Let $\mathbf{X}$ be random variables with marginal distributions $\mathbf{u}$ and copula density $c(\mathbf{u})$. CE of $\mathbf{X}$ is defined as
	\begin{equation}
	H_c(\mathbf{X})=-\int_{\mathbf{u}}{c(\mathbf{u})\log{c(\mathbf{u})}}d\mathbf{u}.
	\end{equation}
\end{definition}

In information theory, MI and entropy are two different concepts \cite{Cover1999}. In \cite{Ma2011}, Ma and Sun proved that they are essentially same -- MI is also a kind of entropy, negative CE, which is stated as follows: 
\begin{theorem}
	\label{thm1}
	MI of random variables is equivalent to negative CE:
	\begin{equation}
	I(\mathbf{X})=-H_c(\mathbf{X}).
	\end{equation}
\end{theorem}
\noindent
The proof of Theorem \ref{thm1} is simple \cite{Ma2011}. There is also an instant corollary (Corollary \ref{c:ce}) on the relationship between information of joint probability density function, marginal density function and copula density function.
\begin{corollary}
	\label{c:ce}
	\begin{equation}
	H(\mathbf{X})=\sum_{i}{H(X_i)}+H_c(\mathbf{X}).
	\end{equation}
\end{corollary}
The above results cast insight into the relationship between entropy, MI, and copula through CE, and therefore build a bridge between information theory and copula theory. CE itself provides a mathematical theory of statistical independence measure.

\subsubsection{Estimation}
\label{s:est}
It has been widely considered that estimating MI is notoriously difficult. Under the blessing of Theorem \ref{thm1}, Ma and Sun \cite{Ma2011} proposed a simple and elegant non-parametric method for estimating CE (MI) from data which comprises of only two steps\footnote{The \textsf{R} package \textsf{copent} for estimating CE is available on CRAN and also on GitHub at \url{https://github.com/majianthu/copent}.}:
\begin{enumerate}
	\item Estimating Empirical Copula Density (ECD);
	\item Estimating CE.
\end{enumerate}

For Step 1, if given data samples $\{\mathbf{x}_1,\ldots,\mathbf{x}_T\}$ i.i.d. generated from random variables $\mathbf{X}=\{x_1,\ldots,x_N\}^T$, one can easily estimate ECD as follows:
\begin{equation}
F_i(x_i)=\frac{1}{T}\sum_{t=1}^{T}{\chi(\mathbf{x}_{t}^{i}\leq x_i)},
\end{equation}
where $i=1,\ldots,N$ and $\chi$ represents for indicator function. Let $\mathbf{u}=[F_1,\ldots,F_N]$, and then one can derive a new samples set $\{\mathbf{u}_1,\ldots,\mathbf{u}_T\}$ as data from ECD $c(\mathbf{u})$. In practice, Step 1 can be easily implemented non-parametrically with rank statistic.

Once ECD is estimated, Step 2 is essentially a problem of entropy estimation which has been contributed with many existing methods. Among them, the kNN method \cite{Kraskov2004} was suggested in \cite{Ma2011}. With rank statistic and the kNN method, one can derive a non-parametric method of estimating CE, which can be applied to any situation without any assumption on the underlying system.

\subsection{Predictive Models}
In this paper, two types of ML algorithms, i.e. Support Vector Machine (SVM) and Random Forests (RF), are selected among many others for building predictive models since they are the most typical and popular methods and are most widely-used in different nonlinear prediction tasks.

SVM is a popular ML method that learns complex relationship from data \cite{Smola2004}. Theoretically, SVM can learn the model with simple model complexity and meanwhile do not compromise on predictive ability, due to the max-margin principle. The learning of SVM model is formulated as an optimization problem \cite{Smola2004}, which can be solved by quadratic programming techniques after transformed to its dual form. SVM has its nonlinear version with kernel tricks. The final SVM model is represented as
\begin{equation}
	f(x) = \sum_{i}{v_i k(x,x_i)+b}
\end{equation}
where $x_i$ represents support vector, and $k(\cdot,\cdot)$ represents kernel function.

RF is another widely-used machine learning algorithms developed by Leo Breiman \cite{Breiman2001}. It learn a model from data by ensembling a group of decision trees. It enjoys a good ability of generalization and model flexibility compared with other machine learning algorithms. Another merits of RF is that feature importance can be derived from the leart models, which make the learning results interpretable.

\section{Experiments and Results}
\label{s:exp}

\subsection{The SDSS Quasar Data}
\begin{figure}
	\centering
	\includegraphics[width=\textwidth]{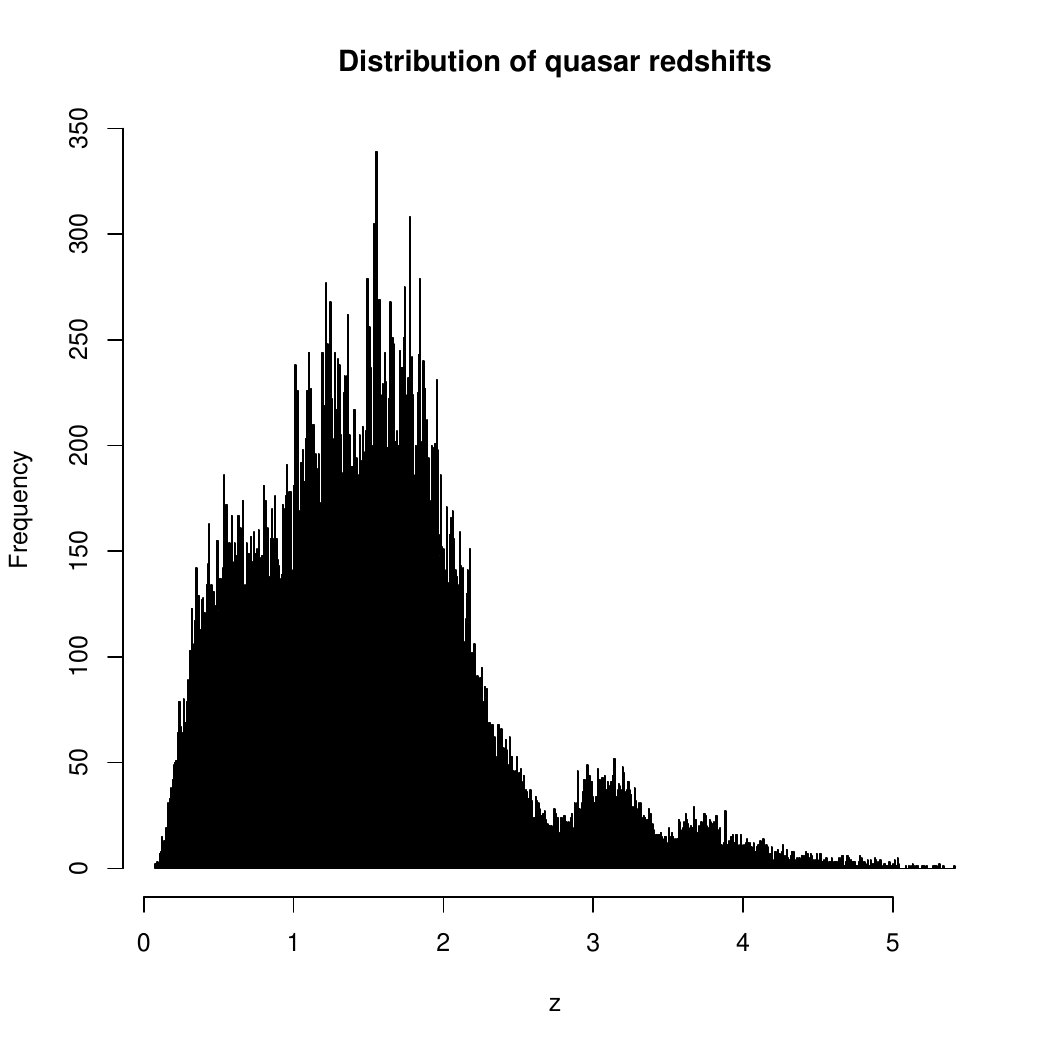}
	\caption{Distribution of redshift z in SDSS quasar dataset.}
	\label{fig:z}
\end{figure}

The data used in this paper is the fourth edition of the Sloan Digital Sky Survey (SDSS) Quasar Catalog \cite{Schneider2007}, which is available in the \textbf{R} package \texttt{astrodatR} \cite{Feigelson2012}. It contains 77,429 objects. The catalog covers about 5740 $deg^2$ area. The quasar redshifts range from 0.08 to 5.41 (median = 1.48). 891 quasars in the catalog has redshifts greater than 4.0, among whom 36 are greater than 5.0. The distribution of quasar redshifts in the SDSS quasar data is shown in Figure \ref{fig:z}.  Each object in the catalog has five-band (ugriz) CCD-based photometry and a luminosity magnitudes. It also contains the factors that indicate whether the objects were detected by NRAO FIRST survey and the ROSAT All-Sky Survey (RASS), but these two factors are not used in our experiments.

\subsection{Experiments}
In the experiment, we want to build models that can predict quasar redshifts from photometric factors, including luminosity and five-band brightness. The key problem is to find which factors are useful for prediction. So we first use CE to measure the statistical dependence between redshifts and photometric factors and select those with large CEs to build the predictive models. Since CE is model-free, it is a good choice for measuring such nonlinear relationships without any assumptions.

For the predictive models, we chose two machine learning algorithms, SVM and random forest (RF). These two algorithms can be used for tackling nonlinear problems and have shown better performance than other machine learning methods \cite{Fernandez-Delgado2014}. Additionally, with RF, we can calculate feature importance from the learned models, which can be a contrast for the CE method.

In the experiment, we separated the whole dateset into training set (the first 5000 samples) and test set (the remaining 72,429 samples). We first used the training set to build the SVM and RF models, and then evaluated the built models on both the training set and the test set. To check whether variable selection can improve the performance of the models, we built two models for each type of model: one with all the factors and the other with the selected factor. So, we will run 8 experiment for SVM and RF on two datasets with two variable sets.

The performance of the SVM and RF models will be measured with mean average error (MAE). Since astronomers are interested in high quasar redshift \cite{Luken2023,Warren1987}, we will also measure the performance of the models on the 851 samples with high redshift ($z>4.0$). We will study whether variable selection with CE can improve the performance of the predictive models on high redshifts.

In the experiment, the \textbf{R} package \texttt{copent} \cite{Ma2021} is used for the implementation of the method for estimating CE from data, and the \textbf{R} package \texttt{e1071} \cite{Chang2011} and \texttt{randomForest} are used as the implementation of the SVM and RF algorithms respectively. The default hyperparameters are adopted in the experiments.


\subsection{Results}
\begin{figure}
	\centering
	\includegraphics[width=\textwidth]{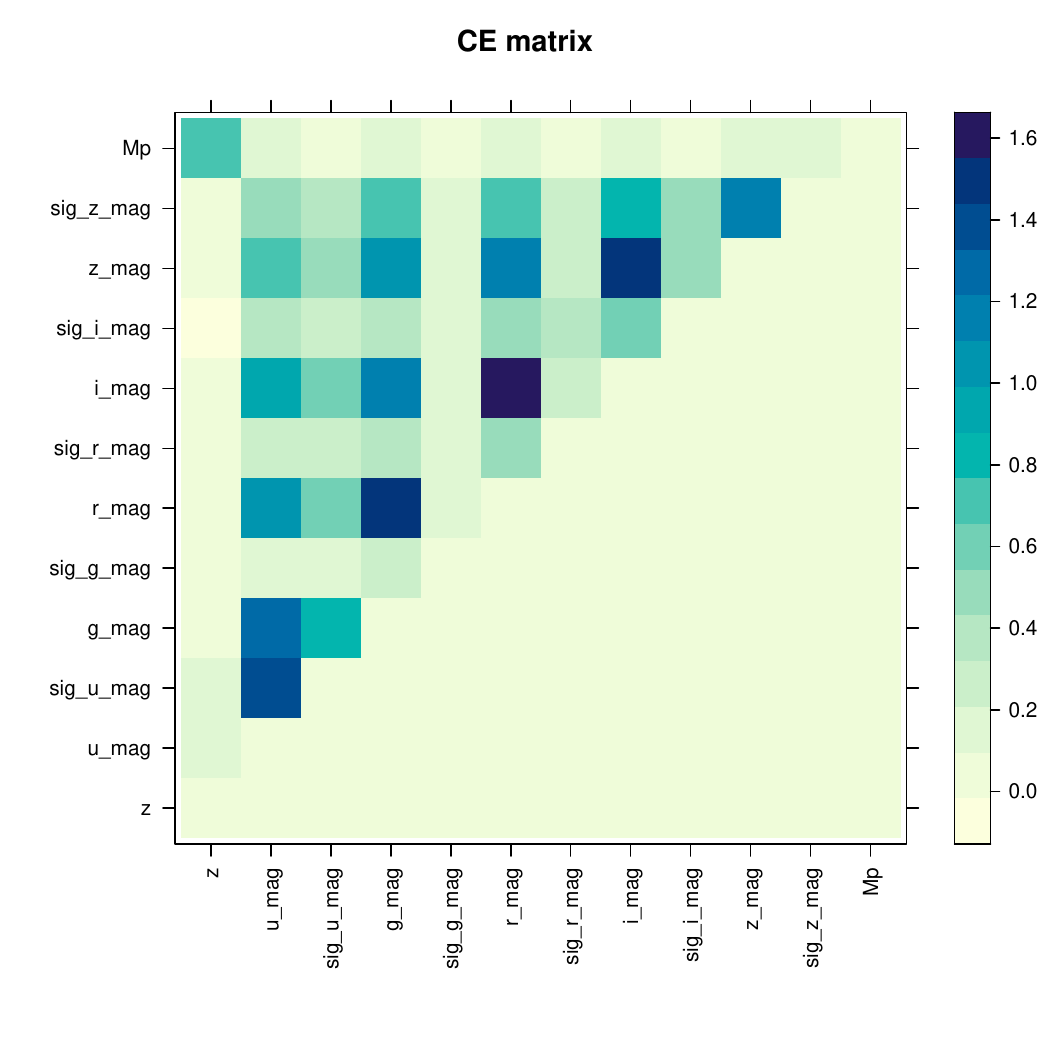}
	\caption{Correlation matrix between measurement of quasars in SDSS quasars dataset.}
	\label{fig:corrmat}
\end{figure}

We estimated the correlation matrix between the factors in the experiments, as shown in Figure \ref{fig:corrmat}. It can be learned from it that the correlations between the photometric factors are strong, especially between the brightness magnitudes of the five-bands.

\begin{figure}
	\centering
	\includegraphics[width=\textwidth]{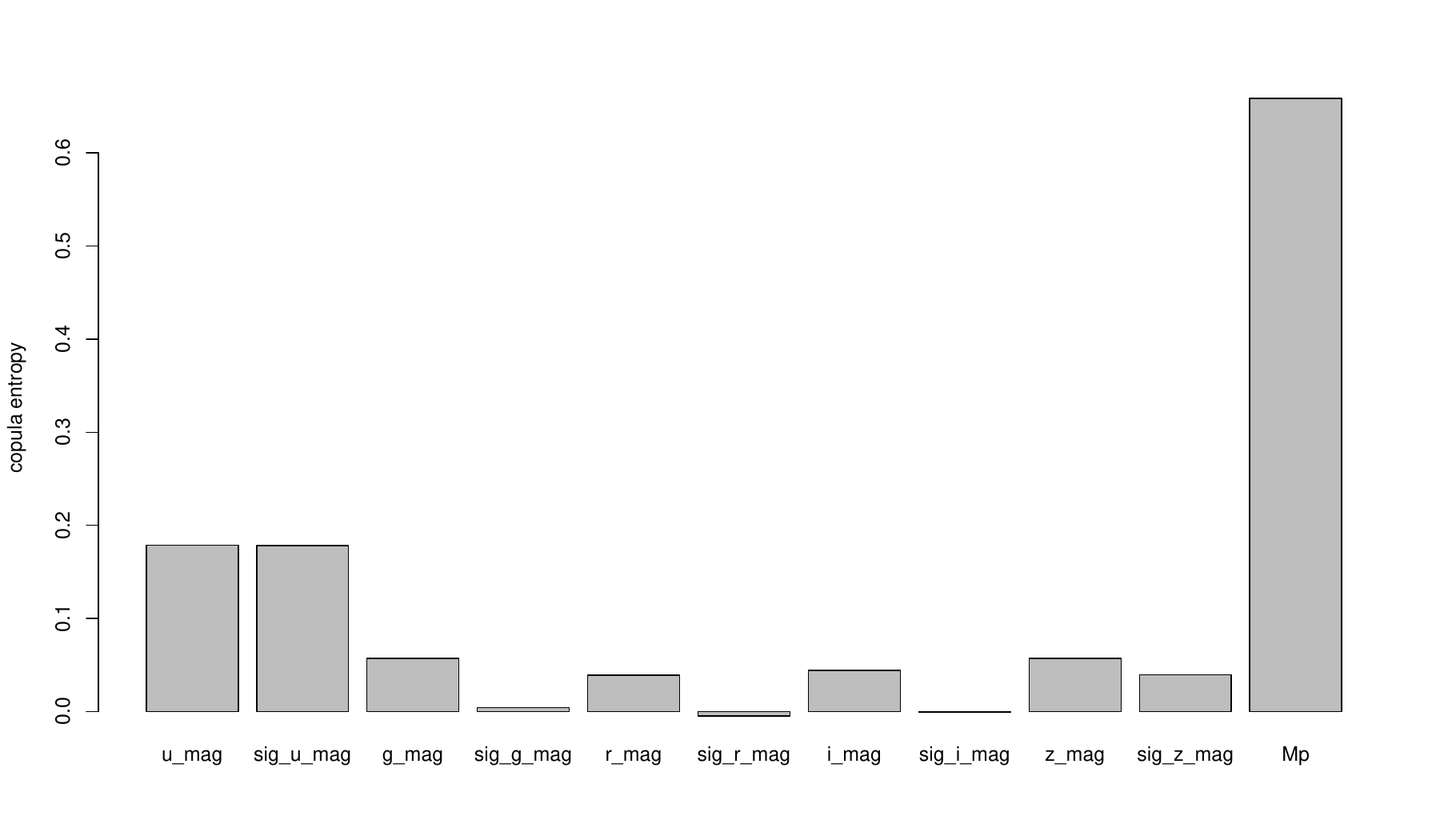}
	\caption{CEs between the redshift z and other measurements in SDSS quasars data.}
	\label{fig:ce}
\end{figure}

The negative CEs between the redshift factor and the other factors were estimated from the data, as shown in Figure \ref{fig:ce}. It can be learned from it that the luminosity magnitude has the strongest dependence with the redshift ($CE_{Mp}$=-0.659), followed by the brightness in the u (ultraviolet) band in  magnitudes with standard deviation ($CE_{u\_mag}$=-0.177, $CE_{sig\_u\_mag}$=-0.178). The brightness in the g (green), r (red), i (further red), and z (further red) band are also associated with strong CEs while the factors of the standard deviation of brightness magnitudes of the five bands have small CE values. So in the next phase of model learning, 7 factors, including luminosity magnitude, the brightness magnitude of the five bands and the standard deviation of the brightness magnitude of the u band, are selected as the inputs of the predictive models.

\begin{figure}
	\centering
	\includegraphics[width=\textwidth]{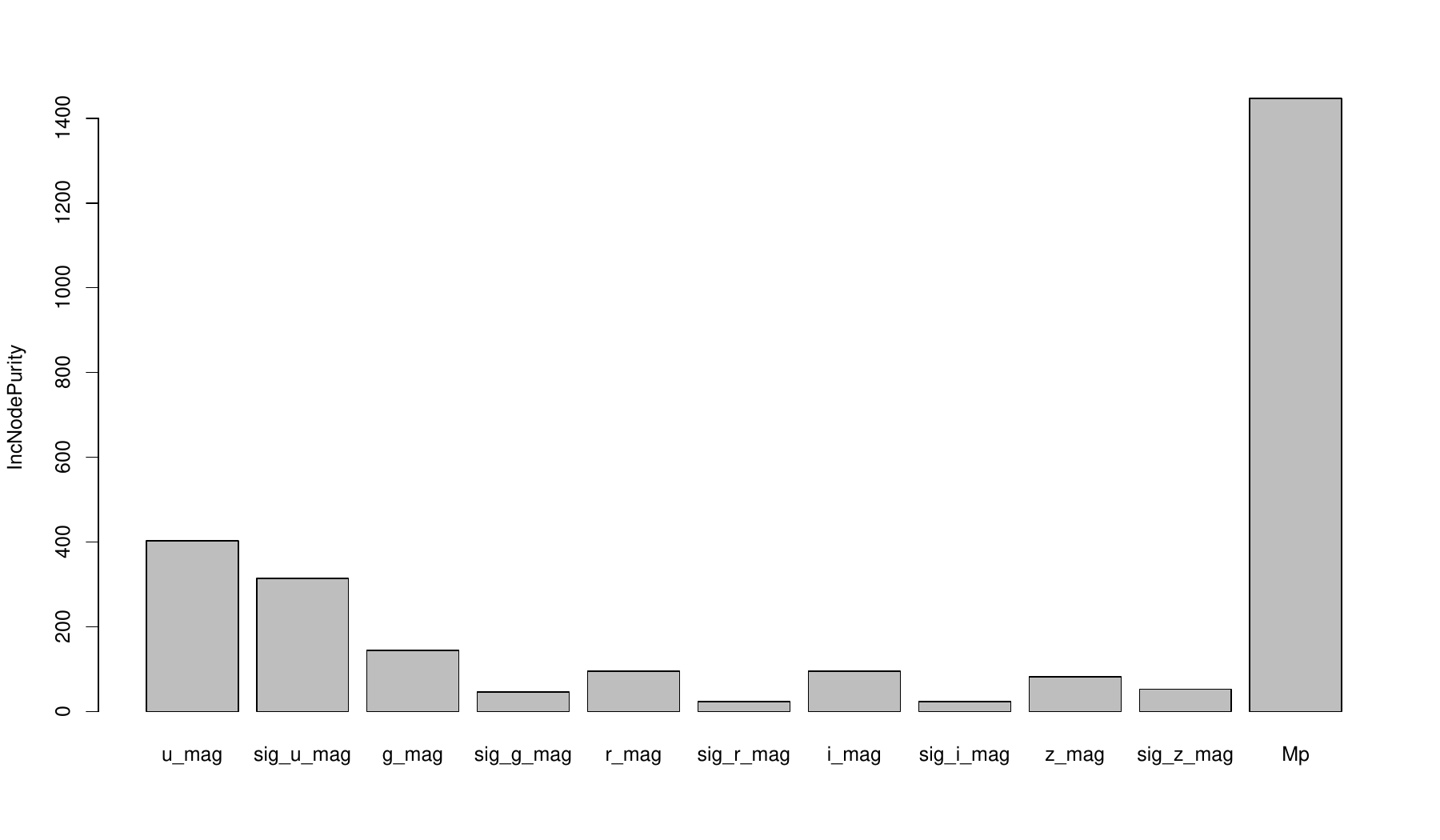}
	\caption{Feature importance in the random forests model for predicting redshift z in the SDSS quasar data.}
	\label{fig:featureimportance}
\end{figure}

As contrast, the feature importance of the factors were also derived from the RF model learned from the training set, as shown in Figure \ref{fig:featureimportance}. It can be learned that the relative importance of the factors presented by CE and RF are very similar.

\begin{figure}
	\centering
	\includegraphics[width=\textwidth]{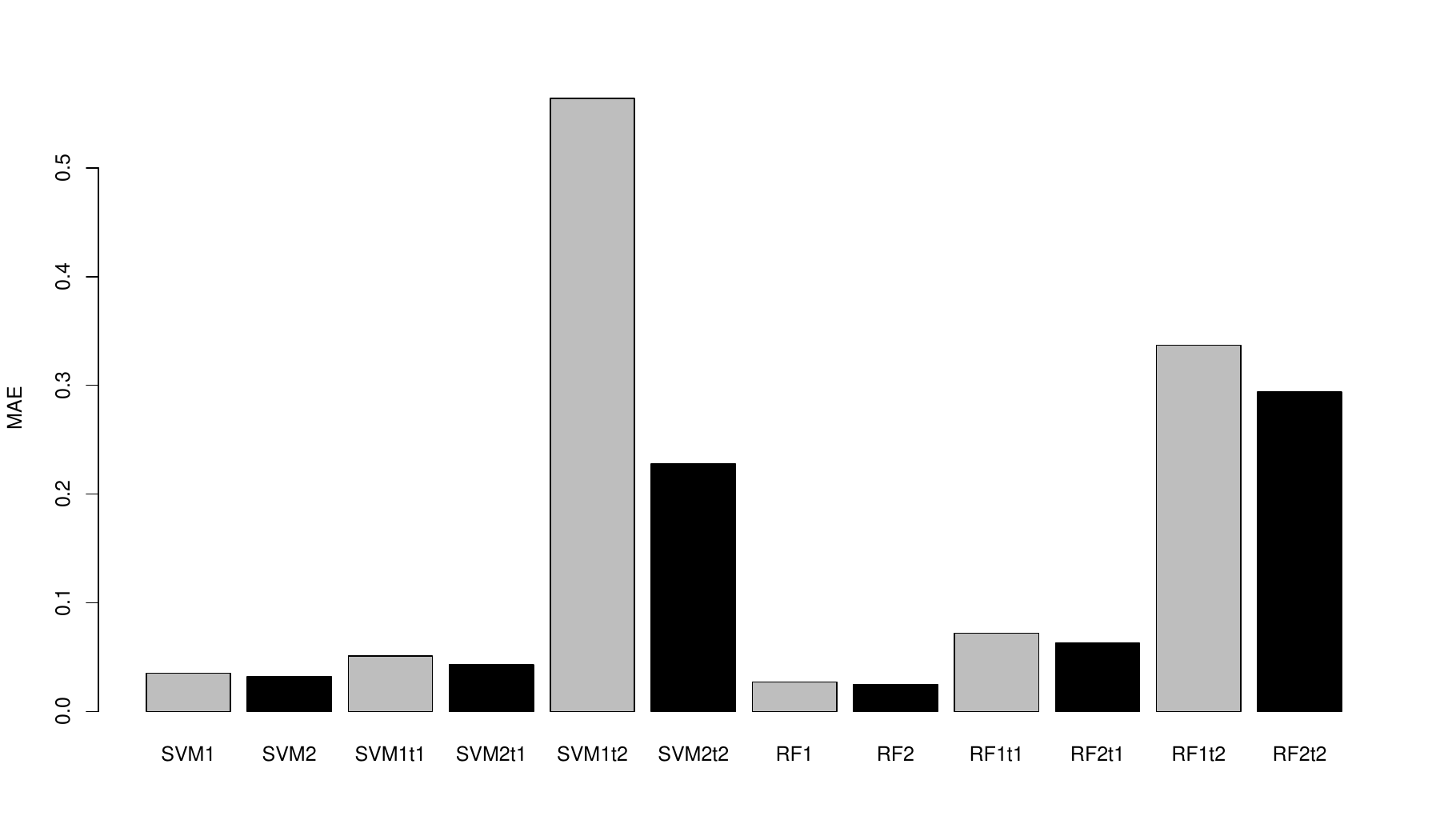}
	\caption{MAEs by SVMs and random forests for predicting redshift z in the SDSS quasar data. `*1' for on the training data, `*2' for on the training data with the selected variables, `*1t1' for on the test data, `*2t1' for on the test data with the selected variables, `*1t2' for on the test data with z $>$ 4.0, `*2t2' for on the test data with z $>$ 4.0 with the selected variables.}
	\label{fig:mae}
\end{figure}

The performance of the SVM and RF models learned from the training set and test set with all the factors and the factors selected with CE were measured by MAE, as shown in Figure \ref{fig:mae}. The MAEs of these models on the test data with redshifts larger than 4.0 are also presented in Figure \ref{fig:mae}. 

It can be learned from Figure \ref{fig:mae} and Table \ref{table:mae} that the performance of the models with the selected factors are always better than that of with all the factors in all the cases. It means that variable selection with CE can improve the performance of the predictive models in all the cases. It can also be learned from Table \ref{table:mae} that the SVM models on the test set with the selected factors present the best performance results in terms of MAE for both the case of the test set as a whole (MAE=0.042) and the case of the test set with high redshift (MAE=0.228). 

\begin{table}
	\centering
	\caption{Performance in terms of MAE of the models learned from the quasar data.}
	\begin{tabular}{l|c|c|c|c}
		\toprule
		Model&factors&training set&test set&test set ($z>0.4$)\\
		\midrule
		SVM&all&0.035&0.051&0.564 \\
		SVM&selected&0.032&\textbf{0.042}&\textbf{0.228} \\
		\hline
		RF&all&0.027&0.072&0.337 \\
		RF&selected&\textbf{0.025}&0.063&0.294\\
		\bottomrule
	\end{tabular}
	\label{table:mae}
\end{table}

The performance of the SVM models is comparable to the counterpart of the RF models in all the comparisons. The MAEs on the test data with high redshift are much larger than that of on the test data as a whole. When comparing SVM and RF on the test data with high redshift, we found that the MAE of SVM is larger than that of RF when all the factors were used while the MAE of SVM is smaller than that of RF when the selected factors were used. The MAE of SVM on the test set with high redshift is improved from 0.564 to 0.228 with CE-based variable selection.

\begin{figure}
	\centering
	\includegraphics[width=\textwidth]{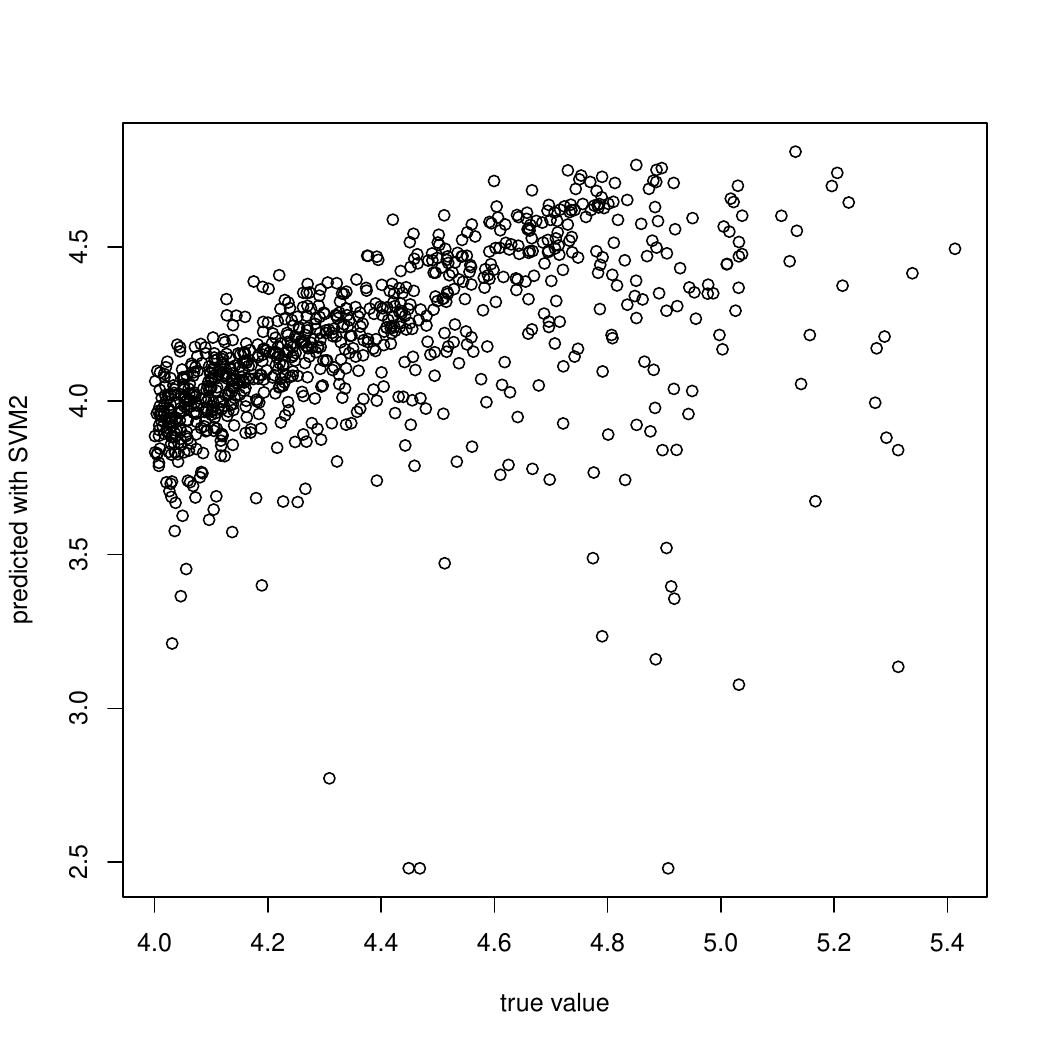}
	\caption{Predicted redshifts ($>4.0$) by SVM with the selected variables.}
	\label{fig:pred2}
\end{figure}

\begin{figure}
	\centering
	\includegraphics[width=\textwidth]{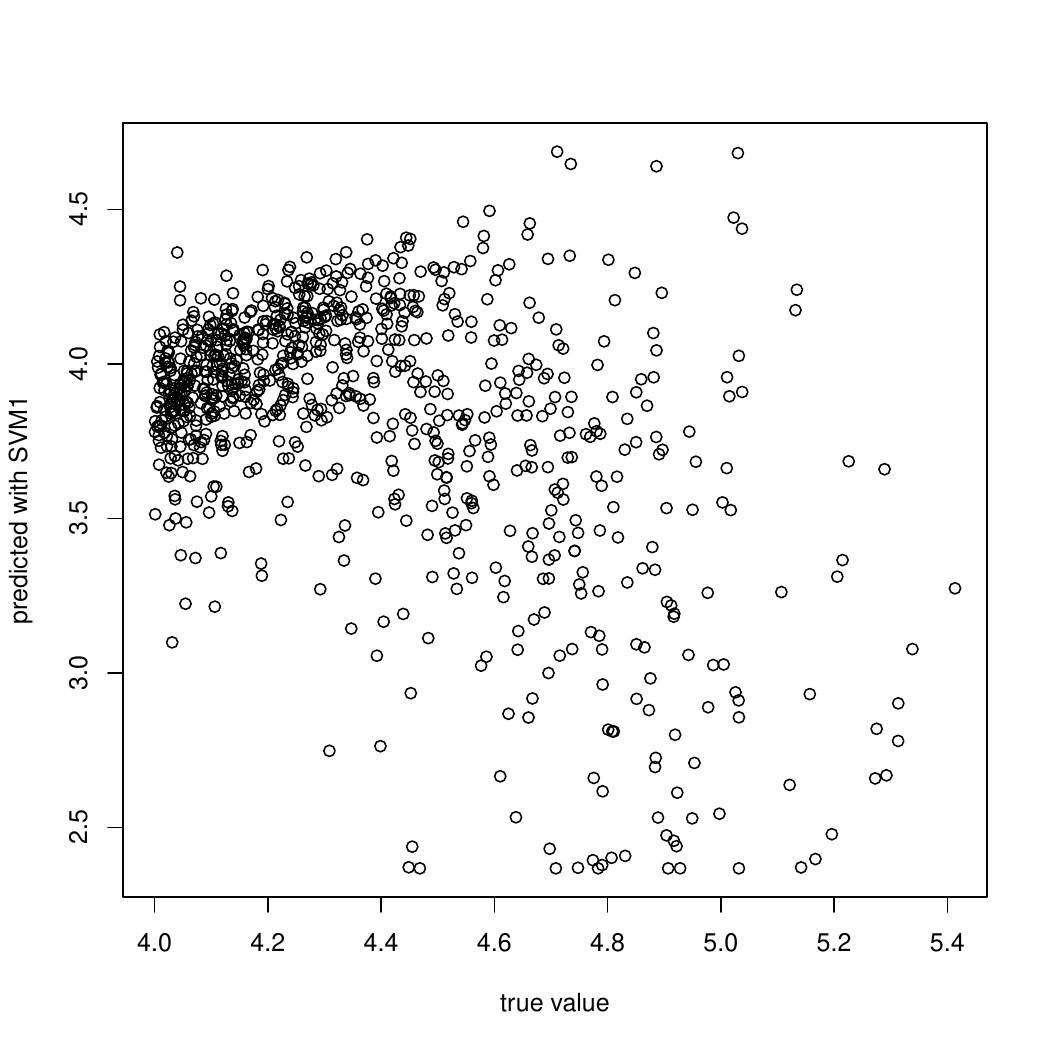}
	\caption{Predicted redshifts ($>4.0$) by SVM with all the variables.}
	\label{fig:pred1}
\end{figure}

We also plot the prediction results of the SVM models to study how the performance of the models were improved by CE. The prediction results of the SVM models for the cases with and without variable selection are shown in Figure \ref{fig:pred2} and \ref{fig:pred1} respectively. It can be learned from the two figures that the accuracy of the predictions on the samples with high redshifts were clearly improved with CE-based variable selection. Particularly, the predictions on the samples with high redshifts are tended to be smaller than the true values by the SVM models learned from all the factors while the predictions by the SVM models learned from the selected factors are much closer to the true value.

\section{Discussion}
\label{sec:discussion}
In the experiments we estimate the CEs between photometric measurements and redshifts as the importance of these factors. As contrast, RF, another commonly used method for deriving feature importance in the machine learning community, is also used. The two results on feature importance are shown in Figure \ref{fig:ce} and Figure \ref{fig:featureimportance}. It can be easily learned that the two results are very similar up to scale. The difference between them is that the results in Figure \ref{fig:ce} is supported with the rigorous CE theory and therefore more accurate.

Two machine learning algorithms, SVM and RF, were used for predicting redshifts in the experiments. It is well known that machine learning models tend to be `black-box' and uninterpretable. RF models can be naively explained according to the structures of the decision trees. Here, the SVM models is also interpretable because the relationships between the inputs and output of the models are built according to the correlations measured with CEs estimated from data. Since CE is model-free and universally applicable, it can also be applied to other problems in astrophysics and cosmology.

\section{Conclusions}
\label{sec:con}
In this paper we propose to apply copula entropy (CE) to photometric redshifts. CE is used to measure the correlations between photometric measurements and redshifts and then the measurements associated with high CEs are selected for predicting redshifts. We verified the proposed method on the SDSS quasar data. Experimental results show that the accuracy of photometric redshifts is improved with the selected measurements compared to the results with all the measurements used in the experiments, especially for the samples with high redshifts. The measurements selected with CE include luminosity magnitude, the brightness in ultraviolet band with standard deviation, and the brightness of the other four bands. Since CE is a rigorously defined mathematical concept, the models such derived is interpretable. In this research, we use the SDSS quasar data (DR5) due to availability. In the future, we expect to apply the method to more latest data, such as the SDSS DR14 \cite{ParisIsabelle2018}.

\bibliographystyle{unsrt}
\bibliography{quasar}

\end{document}